# Vehicle Trajectory Prediction based Predictive Collision Risk Assessment for Autonomous Driving in Highway Scenarios

Dejian Meng, Wei Xiao, Lijun Zhang, Zhuang Zhang, and Zihao Liu

*Abstract*—For driving safely and efficiently in highway scenarios, autonomous vehicles (AVs) must be able to predict future behaviors of surrounding object vehicles (OVs), and assess collision risk accurately for reasonable decision-making. Aiming at autonomous driving in highway scenarios, a predictive collision risk assessment method based on trajectory prediction of OVs is proposed in this paper. Firstly, the vehicle trajectory prediction is formulated as a sequence generation task with long short-term memory (LSTM) encoder-decoder framework. Convolutional social pooling (CSP) and graph attention network (GAN) are adopted for extracting local spatial vehicle interactions and distant spatial vehicle interactions, respectively. Then, two basic risk metrics, time-to-collision (TTC) and minimal distance margin (MDM), are calculated between the predicted trajectory of OV and the candidate trajectory of AV. Consequently, a time-continuous risk function is constructed with temporal and spatial risk metrics. Finally, the vehicle trajectory prediction model CSP-GAN-LSTM is evaluated on two public highway datasets. The quantitative results indicate that the proposed CSP-GAN-LSTM model outperforms the existing state-of-the-art (SOTA) methods in terms of position prediction accuracy. Besides, simulation results in typical highway scenarios further validate the feasibility and effectiveness of the proposed predictive collision risk assessment method.

## I. INTRODUCTION

*A. Motivation*

THE frequent road traffic accidents and traffic jams make the autonomous driving technologies increasingly noticed. The large-scale competitions in the field of autonomous driving have greatly contributed to the research and development of autonomous driving-related technologies by automotive companies and research institutes [1-3]. The simple road structure and the limited driving behaviors under normal conditions make autonomous highway driving the most likely scenario for realistic autonomous driving [4].

For driving safely and efficiently in highway scenarios, AVs must accurately assess situational risk to make reasonable decisions. Situation risk assessment involves two essential steps: situation evolution prediction and collision risk assessment [5]. Situation evolution prediction means that the AV predicts the future behavior, as well as the trajectory of the surrounding dynamic traffic participants, based on the historical track of the local traffic context. Based on situational evolution prediction, the risk assessment module selects appropriate risk indicators and assesses the collision risk between the AV and OVs.

Various types of vehicle collision conflicts have been studied thoroughly and extensively in previous studies, and various collision risk assessment methods have been proposed [6-11]. However, existing studies often use oversimplified assumptions about the future motion of the OVs, such as constant velocity and acceleration models, which fail to capture the complex vehicle interactions in real traffic scenarios. In addition, the previous research on collision risk tends to focus on a specific type of collision conflict, and the risk assessment when multiple types of collision conflicts coexist is not sufficiently considered. Therefore, a predictive collision risk method based on trajectory prediction of OVs is proposed in this work. Taking into account vehicle interactions, the future trajectory of OV is predicted with a data-driven approach, which includes a multi-channel-based LSTM encoder-decoder framework. Then, basic risk metrics are combined for constructing a time-continuous predictive risk function, which supports decision-making for AVs in multi-lane highway scenarios.

*B. Contribution*

Based on the trajectory prediction of OVs, this work proposed a predictive collision risk assessment method for the AV. The main contribution of this work is as follows: (1) We present a three-channel-based LSTM encoder-decoder framework for trajectory prediction of OV. The historical temporal motion characteristic of the OV is extracted in channel 1. Then, the convolution social pooling and graphic attention network are used for modeling local and global vehicle spatial interactions in channel 2 and channel 3, respectively. (2) Based on future motion prediction of OVs, a predictive collision risk assessment suitable for multi-lane highway scenarios is proposed by combining basic temporal and spatial risk metrics. (3) The proposed vehicle trajectory prediction model CSP-

This work was supported by the National Key Research and Development Program of China under Grant No. 2016YFB0100901. (*Corresponding author: Lijun Zhang*.). Dejian Meng and Wei Xiao contributed equally to this work.

D. Meng, W. Xiao, L. Zhang, Z. Zhang, and Z. Liu are with the School of Automotive Studies, Tongji University, Shanghai 201804, China (e-mail: mengdejian@tongji.edu.cn; 1410807@tongji.edu.cn; tjedu_zhanglijun @tongji.edu.cn; zhangzhuang@tongji.edu.cn; 2033544@tongji.edu.cn).

GAN-LSTM outperforms the existing state-of-the-art methods on two public highway datasets in terms of position prediction accuracy.

*C. Paper Organization*

The remainder of the paper is organized as follows. Related works about vehicle trajectory prediction and collision risk assessment are presented in Section II. In section III, the overall framework of the proposed predictive collision risk assessment is briefly introduced. Section IV presents the three-channel-based LSTM encoder-decoder framework for trajectory prediction of the OV. The collision risk assessment between the AV and the OVs is conducted in Section V. After that, Section VI involves a quantitative evaluation for the vehicle trajectory prediction model on two public highway datasets and simulation results analysis for predictive collision risk method in typical highway scenarios. Finally, conclusions and future works are presented in section VII.

## II. RELATED WORKS

This chapter introduces the relevant studies in vehicle trajectory prediction and collision risk assessment.

*A. Vehicle Trajectory Prediction*

To ensure safe and efficient driving in complex dynamic traffic scenarios, AVs need to accurately sense the current states of traffic participants in the driving environment, and make predictive inferences about their future driving intentions. Unlike the urban road traffic scenario where many types of traffic participants coexist, the traffic participants in the highway scenario are homogeneous. Besides, there is a strong correlation between the vehicle motion and the road structure, which provides some convenience for the vehicle trajectory prediction in highway scenarios. According to the level of information abstraction and model completeness, vehicle trajectory prediction methods can be divided into three categories: physical-based, behavior-based, and interaction-aware-based methods [7]. Since the interaction-aware-based vehicle trajectory prediction methods have become the current SOTA methods in this field, we only briefly introduce the first two methods, mainly focusing on the interaction-aware-based vehicle trajectory prediction method.

The physics-based approaches treat vehicles as entities controlled by physical laws and use kinematic or dynamical models to predict future motion [12-13]. Since only the current motion state information is utilized, the physics-based vehicle trajectory prediction method is only reliable for a short prediction horizon (less than 1s) [7]. The behavior-based vehicle trajectory prediction approaches treat the trajectory of vehicles on a road network as a series of mutually independent behaviors [14-15]. Behaviors are recognized with machine learning or deep learning methods. Then the future trajectory is predicted with recognized behavior and predefined trajectory prototypes related to each type of behavior. Compare with physics-based approaches, the behavior-based methods have better prediction accuracy. However, the inherent characteristic of vehicle interactions in traffic scenarios is ignored.

The interaction-aware-based vehicle trajectory prediction methods incorporate the interconnection among vehicles. Currently, interaction-aware trajectory prediction based on deep learning frameworks has become common approaches adopted by researchers, which uses a data-driven way to learn the interaction among vehicles. A deep-learning-based vehicle trajectory prediction survey is presented in [16], which summaries the current deep learning-based vehicle trajectory prediction methods from three aspects: input representation, output types, and prediction models. The input representations include four different types of data: the historical track of the OV, the historical track of the OV and its surrounding vehicles (SVs), a simplified bird's eye view, and raw sensor data. The output types involve behavior, single trajectory, multi-modal trajectory, and occupy grid map. The prediction models mainly include recurrent neural networks (RNNs), convolutional neural networks (CNNs) and others (including the combination of RNNs and CNNs, graph neural networks (GNN)).

As the prediction time increases, the correlation between the vehicle trajectory in the prediction horizon and its historical trajectory gradually weakens. Vehicle trajectory in the future is uncertain and multi-modal. The future motion of the OV is influenced by its SVs and the road structure. These factors make the vehicle trajectory prediction task very challenging. Essentially, vehicle trajectory prediction is a time series generation task. The historical track of the OV and its SVs is used as input to predict the future trajectory of the OV. The key to accurate vehicle trajectory prediction lies in complex vehicle spatial-temporal interactions modeling in the local traffic context. Considering the advantages of LSTM in modeling nonlinear temporal relations and CNN in spatial relation extraction, a convolutional social pooling-based LSTM encoder-decoder architecture was proposed for vehicle trajectory prediction in highway scenarios [17]. Following the identical LSTM encoder-decoder framework, a multi-head attention mechanism was adopted modeling distant spatial vehicle interactions between the OV and the SVs [18]. The SVs' influence on the future motion of the OV was represented by attention mechanism. A structural LSTM architecture was proposed in [19], which modeled vehicle interaction by sharing implicit and cellular states in the LSTM network based on radial spatial connection relationships. Vehicle interactions were expressed with graph-based convolutional neural networks in [20]. The traffic graph was adopted to represent the spatial interactions. The CNN was used to extract temporal characteristics. In [21], a weighted dynamic geometric graph was introduced to indicate vehicle interactions. The future trajectory of the OV is predicted with a two-stream-based LSTM encoder-decoder framework. Stream 1 involved encoding historical motion information of the OV. Stream 2 contained the spectrum of the traffic graph.

In this work, we present a three-channel-based LSTM encoder-decoder framework for trajectory prediction of the OV. Channel 1 includes only the historical motion information of the OV. Channel 2 is a convolutional social pooling-based local spatial vehicle interactions module, which involves all local vehicle interactions in the local traffic context around the OV.

Channel 3 extracts distant vehicle interactions between the OV and the SVs. Then, the three-channel features are aggregated and fed into the LSTM decoder to predict the future trajectory of the OV.

*B. Collision Risk Assessment*

Depending on whether the motion uncertainty is considered or not, risk assessment methods can be divided into deterministic risk assessment methods and probabilistic risk assessment methods [22].

Under the assumption of a single behavior of the OV, deterministic risk assessment methods usually contain a simplified vehicle motion prediction model to describe the future motion of the OV. Collision risk can be assessed in terms of the temporal proximity and spatial proximity of possible collision conflict events and the feasibility of collision avoidance behavior. TTC (Time-to-Collision) [23], TH (Time Headway), [24] and PET (Post-Encroachment Time) [25] are the most commonly used time-domain based collision risk metrics, which are widely used for collision risk assessment in car-following scenarios. The most commonly used spatial domain collision risk metrics include MSD (Minimum Safe Distance) [26], PSD (Proportion of Stopping Distance) [27], and SHD (Stopping Headway Distance) [28], etc. Collision avoidance behavior feasibility-based risk metrics include BTN (Brake Threat Number) [9], STN (Steering Threat Number) [9], TTR (Time-to-React) [9], TTB (Time-To-Brake) [9], TTK (Time-To-Kickdown) [9], and TTS (Time-To-Steer) [9], etc.

There are different types of collision conflicts in real traffic scenarios, such as rear-end collisions, head-on collisions, and side-on collisions. Many existing studies focus only on one type of collision conflict, and the proposed single risk indicator is only applicable to a specific scenario. For instance, both TTC and TH are only applicable to rear-end collision risk assessment in the single-lane car following scenarios and are not suitable for collision risk assessment in multi-lane highway scenarios. Using only the TTC indicator tends to underestimate the risk of collision when two vehicles are traveling at similar high speeds with small spacing. Similarly, using only the TH indicator tends to underestimate the collision risk when the preceding vehicle is traveling at a low speed and the following vehicle is traveling at a high speed. Some existing studies deal with collision risk assessment when different types of traffic conflicts coexist [5, 10, 29-30]. The literature [10] introduced the lateral position influence factor for modifying the TH and TTC indicators. The modified TH and TTC metrics were combined for collision risk assessment in multi-lane highway scenarios. Similarly, the collision risk in multi-lane was evaluated with TTC, TH, and the originally proposed TTF (Time-To-Front) [5] metric, which was decided by the lateral velocity and the lateral offset. A single risk indicator is only applicable to collision risk assessment under specific types of traffic conflicts. The combination of multiple risk indicators can more accurately characterize the collision risk under multiple types of traffic conflicts in complex scenarios.

Probabilistic risk assessment methods [22, 31-33] include the vehicle motion uncertainty with Gaussian distribution or random reachable set. The probabilistic collision risk assessment is implemented with Monte Carlo simulation. In literature [31], Monte Carlo simulation and Markov chain are used for predicting the road grid occupation probability of the OV. The collision risk was obtained by calculating the probability that the ego vehicle and the OV were located in the same grid. In literature [22], the behavior probability and the corresponding future trajectory of the OV were predicted with an interactive multi-model. The collision risk between the ego vehicle and the OV was computed based on TTC metrics and the probability distribution of OV's trajectories.

Although various types of vehicle collision conflicts have been studied thoroughly and extensively in previous studies, various collision risk assessment methods have been proposed. However, existing studies often use oversimplified assumptions about the future motion of the OVs, which fail to capture the complex vehicle interactions in real traffic scenarios. Therefore, in this paper, based on accurately predicting the future trajectory of the OV around the AV, a multiple risk metrics-based collision risk assessment method is proposed, which comprehensively evaluates the collision risk when multiple types of traffic conflicts coexist.

## III. OVERALL FRAMEWORK

The proposed predictive collision risk assessment architecture for autonomous driving in multi-lane highway scenarios is shown in Fig. 1. It is assumed that the historical motion state information of the AV and the OVs and road structure information can be provided by the environment perception module. Using the historical motion state sequence of the OV and its SVs as input, the future trajectory of the OV is predicted with a data-driven approach, which involves the three-channel-based LSTM encoder-decoder framework. Then, the candidate trajectories of the AV are described under possible longitudinal and lateral behaviors. Meanwhile, the segmented cubic spline interpolation curve is used to transform the discrete position of the OV generated by the vehicle trajectory prediction module into time-continuous longitudinal and lateral positions in the prediction horizon. Further, based on the candidate trajectories of the AV and the predicted trajectories of the OVs in the prediction horizon, the basic risk metrics TTC and MDM are calculated, respectively. A time-continuous collision risk function is constructed by combing basic TTC and MDM risk metrics and is adopted to assess the collision risk between the AV and the OVs.

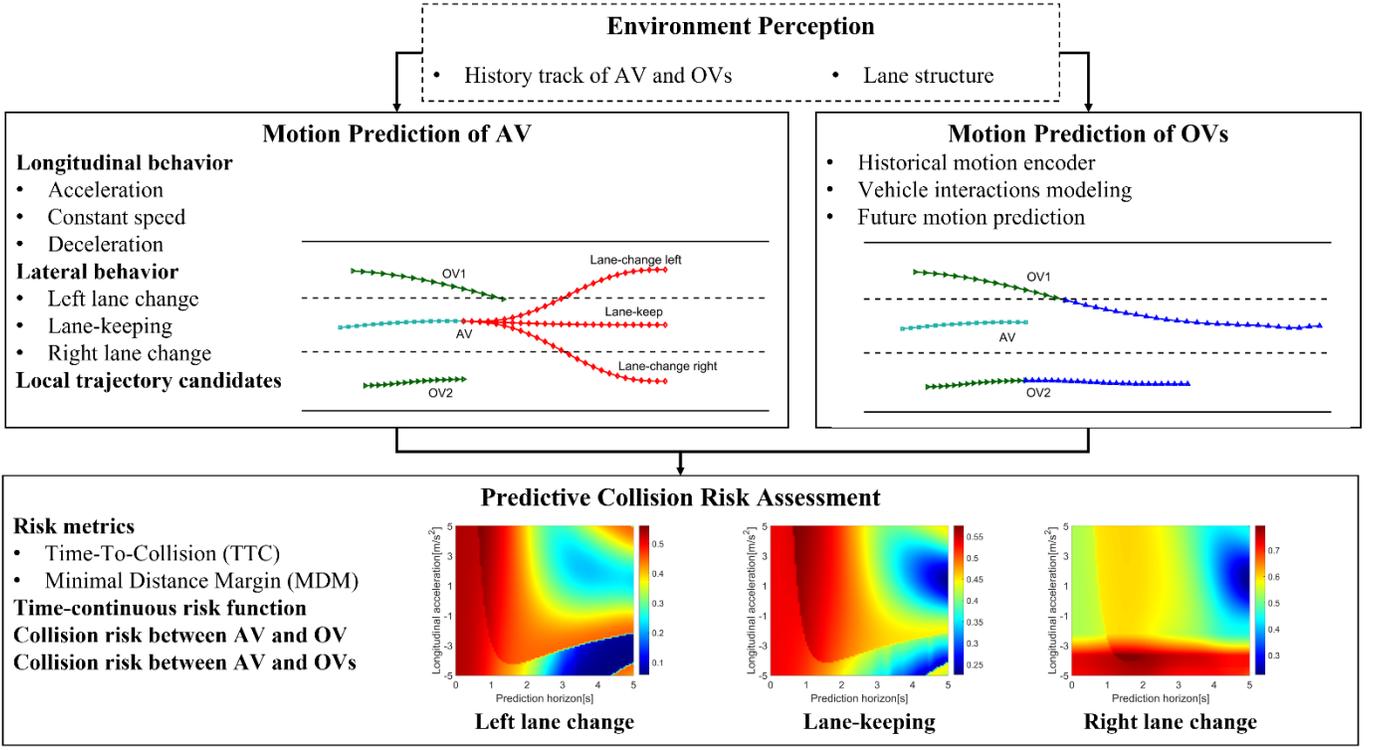

**Fig. 1.** The proposed predictive collision risk assessment framework for highway autonomous driving.

## IV. VEHICLE TRAJECTORY PREDICTION OF THE OV

The trajectory prediction of the OVs around the AV takes the historical track of the OVs and their SVs and lane structure as input. Essentially, the trajectory prediction of the OVs is a typical multiple-input-multiple-output task. Considering the model complexity and computational time, this paper adopts a multiple-input-single-output framework to decompose the original multiple-input-multiple-output task into multiple multiple-input-single-output tasks.

### A. Input and Output

The historical track of the OV and its SVs is used as input for the vehicle trajectory prediction model, which mainly includes the absolute motion of the OV and the SVs in the road curve coordinate system, and the relative motion of the SVs to the OV. The SVs are selected by spatial adjacency, as illustrated in Fig. 2. The local traffic context involves 12 vehicles. The input for the missing SV is replaced with a zero vector.

$$\begin{cases} X = \left[ X_{OV}, X_{SV_1}, \cdots, X_{SV_N} \right] \\ X_{OV} = \left[ x_{OV}(t_0 - t_h), x_{OV}(t_0 - t_h + \Delta t), \cdots, x_{OV}(t_0) \right] \\ X_{SV_i} = \left[ x_{SV_i}(t_0 - t_h), x_{SV_i}(t_0 - t_h + \Delta t), \cdots, x_{SV_i}(t_0) \right], i = lb, \cdots, rff \\ x_{OV}(t) = \begin{bmatrix} x_{OV}(t), y_{OV}(t), \dot{x}_{OV}(t), \\ \dot{y}_{OV}(t), \ddot{x}_{OV}(t), \ddot{y}_{OV}(t) \end{bmatrix}^T, t \in [t_0 - t_h, t_0] \\ x_{SV_i}(t) = \begin{bmatrix} x_{SV_i}(t), y_{SV_i}(t), \dot{x}_{SV_i}(t), \dot{y}_{SV_i}(t), \\ \ddot{x}_{SV_i}(t), \ddot{y}_{SV_i}(t), \Delta x_{SV_i}(t), \Delta y_{SV_i}(t), \\ \Delta \dot{x}_{SV_i}(t), \Delta \dot{y}_{SV_i}(t), \Delta \ddot{x}_{SV_i}(t), \Delta \ddot{y}_{SV_i}(t) \end{bmatrix}^T, t \in [t_0 - t_h, t_0] \end{cases} \quad (1)$$

Where $X$ is the input for the vehicle trajectory model, $X_{OV}$ and $X_{SV_i}$ are historical track of the OV and SV, respectively, $[t_0 - t_h, t_0]$ is the historical horizon, $t_h$ is the length of historical horizon, $\Delta t$ is the time interval, $x_{OV}$ and $x_{SV_i}$ are the motion state vector of the OV and SV, respectively.

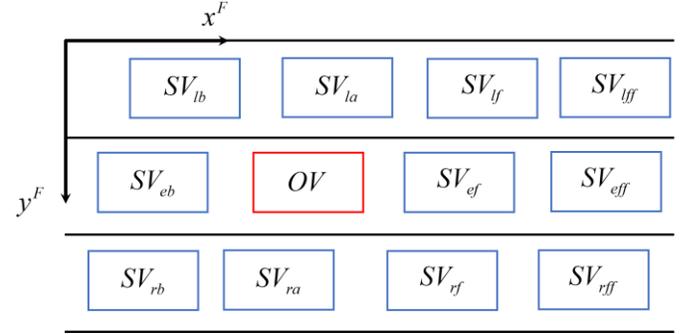

**Fig. 2.** The local traffic context around the OV.

Considering the prediction uncertainty, it is assumed that the predicted position of the OV obeys a binary Gaussian distribution. The output of the vehicle trajectory prediction model is probabilistic distribution parameters of the OV's predicted position.

$$\begin{cases} Y_{OV} = \left[ y_{OV}(t_0 + \Delta t), y_{OV}(t_0 + 2\Delta t), \cdots, y_{OV}(t_0 + t_f) \right] \\ y_{OV}(t) = \left[ x_{OV}(t), y_{OV}(t) \right]^T, t \in [t_0, t_0 + t_f] \\ y_{OV}(t) \sim \mathbb{N}(\boldsymbol{\mu}(t), \boldsymbol{\Sigma}(t)) \\ \boldsymbol{\mu}(t) = \begin{bmatrix} \mu_x(t) \\ \mu_y(t) \end{bmatrix}, \boldsymbol{\Sigma}(t) = \begin{bmatrix} (\sigma_x(t))^2 & \rho_{xy}(t)\sigma_x(t)\sigma_y(t) \\ \rho_{xy}(t)\sigma_x(t)\sigma_y(t) & (\sigma_y(t))^2 \end{bmatrix} \\ \boldsymbol{\Theta} = \left[ \Theta(t_0 + \Delta t), \Theta(t_0 + 2\Delta t), \cdots, \Theta(t_0 + t_f) \right] \\ \Theta(t) = \left[ \mu_x(t), \mu_y(t), \sigma_x(t), \sigma_y(t), \rho_{xy}(t) \right]^T, t \in [t_0, t_0 + t_f] \end{cases} \quad (2)$$

Where $Y_{OV}$ is the predicted trajectory, $y_{ov}$ is the predicted position, $[t_0, t_0+t_f]$ is the prediction horizon, $t_f$ is the length of the prediction horizon, $\mu$ and $\Sigma$ are mean value vector and covariance matrix, respectively, $\Theta$ is the probabilistic distribution parameter related to the predicted trajectory.

*B. Loss Functions*

The vehicle motion prediction is a typical sequence generation task. The commonly used loss functions for model parameters training include the root mean square error (RMSE) [19-20] and negative log-likelihood (NLL) loss functions [17-18]. In this work, the RMSE loss function is adopted for model parameters pre-training, and the NLL loss function is used for formal training.

$$L_{RMSE} = \sqrt{\frac{1}{N_f} \sum_{t=t_0+\Delta t}^{t_0+t_f} \left[ y_{OV,pred}(t) - y_{OV,GT}(t) \right]^2} \quad (3)$$

$$L_{NLL} = -\log\left( \prod_{t=t_0+\Delta t}^{t_0+t_f} P_{\Theta(t)}\left(y_{OV,GT}(t)|X\right) \right)$$
$$= -\sum_{t=t_0+\Delta t}^{t_0+t_f} \log P_{\Theta(t)}\left(y_{OV,GT}(t)|X\right) \quad (4)$$

Where $N_f = t_f/\Delta t$ is the prediction steps, $y_{OV,GT}$ is the true position of the OV, $y_{OV,pred} = [\mu_x, \mu_y]^T$ is the mean vector for the predicted position, $P_{\Theta(t)}(y_{OV,GT}(t)|X)$ is the probability that the true position appears around the predicted mean vector.

*C. Vehicle Trajectory Prediction Model Overall Framework*

The proposed vehicle trajectory prediction model consists of the historical motion encoder module, the vehicle interactions modeling module, and the future motion prediction module, as illustrated in Fig. 3. The historical motion encoder module uses LSTM networks to extract the temporal characteristics of each vehicle in the local traffic context. In the vehicle interactions modeling module, convolutional social pooling and graph attention network are used for extracting local and global vehicle interactions, respectively. Finally, the future motion decoder module predicts the future trajectory of the OV with the context vector fused by the three channels.

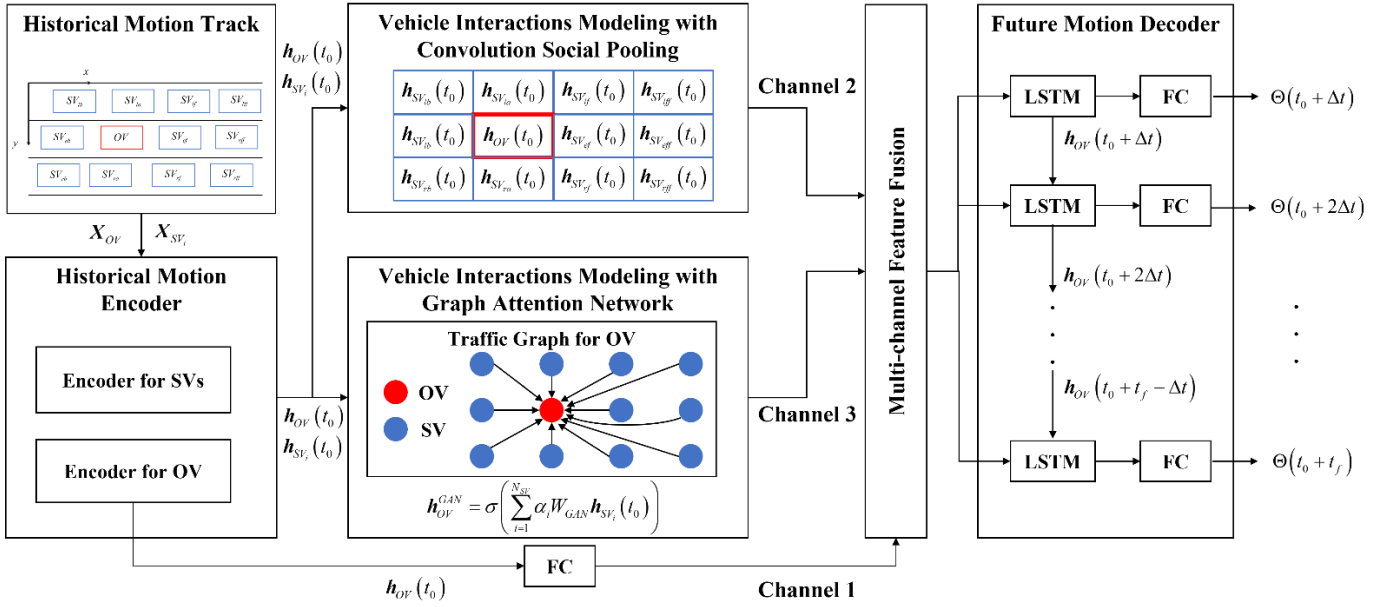

**Fig. 3.** The vehicle trajectory prediction model framework.

*D. Historical Motion Encoder*

In the historical motion encoder module, the fully connected (FC) layers are used to embed the historical motion state of the OV and the SVs, respectively. Then, LSTM networks are adopted to extract the temporal relationships in the historical motion sequences [17-18].

$$\begin{cases} e_{OV}^{emb}(t) = FC_{OV}^{emb}\left(x_{ov}(t); W_{OV}^{emb}\right) \\ h_{OV}^{enc}(t), c_{OV}^{enc}(t) = LSTM_{OV}^{enc}\left(e_{OV}^{emb}(t), h_{OV}^{enc}(t-\Delta t), c_{OV}^{enc}(t-\Delta t); W_{OV}^{enc}\right) \end{cases} \quad (5)$$

$$\begin{cases} e_{SV_i}^{emb}(t) = FC_{SV}^{emb}\left(x_{SV_i}(t); W_{SV}^{emb}\right) \\ h_{SV_i}^{enc}(t), c_{SV_i}^{enc}(t) = LSTM_{SV}^{enc}\left(e_{SV_i}^{emb}(t), h_{SV_i}^{enc}(t-\Delta t), c_{SV_i}^{enc}(t-\Delta t); W_{SV}^{enc}\right) \end{cases} \quad (6)$$

Where $FC_{OV}^{emb}(\cdot)$ and $FC_{SV}^{emb}(\cdot)$ are FC networks, $W_{OV}^{emb}$ and $W_{SV}^{emb}$ are parameters for FC networks, $LSTM_{OV}^{enc}(\cdot)$ and $LSTM_{SV}^{enc}(\cdot)$ are LSTM networks, $W_{OV}^{enc}$ and $W_{SV}^{enc}$ are parameters for LSTM networks, $e_{OV}^{emb}$ and $e_{SV_i}^{emb}$ are embedded motion state vector, $h_{OV}^{enc}$, $c_{OV}^{enc}$, $h_{SV_i}^{enc}$, and $c_{SV_i}^{enc}$ are hidden state vector and output vector in LSTM networks.

*E. Vehicle Interactions Modelling*

The future motion of the OV is related to its historical motion and is also influenced by the interactions among vehicles in the local traffic context. LSTM networks have excellent performance in modeling nonlinear temporal relations. However, LSTM networks are not adequate for modeling spatial interactions. Thus, CNNs [17] and attention mechanisms

[18] are introduced for vehicle spatial interactions modeling. The convolutional pooling operation captures local spatial vehicle interactions. The attention mechanism extracts distant and global spatial vehicle interactions. In this work, we combine the two approaches in modeling vehicle interactions. Instead of using spatial distance-based local traffic context, we adopt a spatial adjacency relationship-based local traffic context for constructing the traffic graph around OV. Then, convolutional social pooling and graph attention network are adopted for vehicle interactions modeling.

*1) Historical Motion Characteristics Extraction for the OV (Channel 1):* The FC layer is used to map the hidden state vector $h_{OV}^{enc}(t_0)$ of the OV into a context vector $c_{ch1}$, which includes the historical temporal motion characteristics of the OV.

$$c_{ch1} = FC_{OV}^{dyn}\left(h_{OV}(t_0);W_{OV}^{dyn}\right) \quad (7)$$

Where $FC_{OV}^{dyn}(\cdot)$ is the FC layer, $W_{OV}^{dyn}$ is the parameter for the FC layer.

*2) Vehicle Interactions Modelling with Convolutional Social Pooling (Channel 2):* Based on the current spatial configuration between the OV and SVs, the hidden state vector $h_{OV}^{enc}(t_0)$ and $h_{SV_i}^{enc}(t_0)$ are assembled into a social tensor $H$. The social tensor $H$ involves both the spatial configuration of the local traffic context and the temporal characteristic of each vehicle.

$$H = \begin{bmatrix} h_{SV_{lff}}^{enc}(t_0) & h_{SV_{eff}}^{enc}(t_0) & h_{SV_{rff}}^{enc}(t_0) \\ h_{SV_{lf}}^{enc}(t_0) & h_{SV_{ef}}^{enc}(t_0) & h_{SV_{rf}}^{enc}(t_0) \\ h_{SV_{la}}^{enc}(t_0) & h_{OV}^{enc}(t_0) & h_{SV_{ra}}^{enc}(t_0) \\ h_{SV_{lb}}^{enc}(t_0) & h_{SV_{eb}}^{enc}(t_0) & h_{SV_{rb}}^{enc}(t_0) \end{bmatrix} \quad (8)$$

Following the processing way in the literature [17], two successive spatial convolutional operations and one max-pooling operation are imposed on the social tensor $H$. The convolutional social pooling extracts local spatial interactions among vehicles in the local traffic context around the OV.

$$\begin{cases} H^{(1)} = LeakyReLU\left(Conv2D^{(1)}\left(H;W^{Conv2D^{(1)}}\right)\right) \\ H^{(2)} = LeakyReLU\left(Conv2D^{(2)}\left(H^{(1)};W^{Conv2D^{(2)}}\right)\right) \\ c_{ch2} = MaxPooling2D\left(H^{(2)}\right) \end{cases} \quad (9)$$

Where $Conv2D^{(1)}(\cdot)$ and $Conv2D^{(2)}(\cdot)$ are two-dimensional CNNs, the kernel sizes are $2\times2$ and $1\times2$, respectively, $W^{Conv2D^{(1)}}$ and $W^{Conv2D^{(2)}}$ are parameters for CNNs, $LeakyReLU(\cdot)$ is activation function, $MaxPooling2D(\cdot)$ is the max-pooling layers, the kernel size is $2\times1$, $c_{ch2}$ is the context vector generated by channel 2, which includes local vehicle interactions.

*3) Vehicle Interactions Modelling with Graph Attention Network (Channel 3):* The future motion of the OV is influenced by its SVs. Graph attention network is adopted to extract global spatial interactions between the OV and the SVs.

The SVs' influence on the OV is constructed as a directed graph. The node in the graph represents each vehicle, and the edge indicates interactions between the OV and the SV. Based on the graph attention network, the global vehicle interactions are introduced.

$$\begin{cases} c_{ch3} = h_{OV}^{GAN} = \sigma\left(\sum_{i=1}^{N_{SV}} \alpha_i W_{GAN} h_{SV_i}(t_0)\right) \\ \alpha_i = \text{softmax}(e_i) = \dfrac{\exp(e_i)}{\sum_{i=1}^{N_{SV}} \exp(e_i)} \\ e_i = Score\left(W_{GAN} h_{SV_i}(t_0), W_{GAN} h_{OV}(t_0)\right) \end{cases} \quad (10)$$

Where $h_{OV}^{GAN}$ is output of the GAN, $e_i$ is the influence factor of the SV, $\alpha_i$ is normalized weighting coefficient, $Score(\cdot)$ is the score function, $\text{Softmax}(\cdot)$ is normalized function, $W_{GAN}$ is the parameters for FC layer used for feature projection, $\sigma(\cdot)$ is activation function, $c_{ch3}$ is the context vector generated by channel 3.

*4) Multi-channel Feature Fusion:* The context vectors generated by three channels are fused into a complete context vector.

$$c_{cont} = c_{ch1} \oplus c_{ch2} \oplus c_{ch3} \quad (11)$$

Where $c_{cont}$ is the complete context vector, $\oplus$ is vector concatenation operation.

*F. Future Motion Decoder*

In the future motion prediction module, the context vector $c_{cont}$ is used as input for the LSTM decoder, which involves historical motion information of the OV and the vehicle interactions in the local traffic context. The future motion prediction module outputs probabilistic distribution parameters of the predicted trajectory.

$$h_{OV}^{dec}(t), c_{OV}^{dec}(t) = LSTM_{OV}^{dec}\left(h_{OV}^{dec}(t-\Delta t), c_{OV}^{dec}(t-\Delta t), c_{cont}; W_{OV}^{dec}\right) \quad (12)$$

$$\Theta(t) = FC_{OV}^{out}\left(c_{OV}^{dec}(t); W_{OV}^{out}\right) \quad (13)$$

Where $LSTM_{OV}^{dec}(\cdot)$ is LSTM network for decoding future motion of the OV, $W_{OV}^{dec}$ is parameter for the LSTM network, $FC_{OV}^{out}(\cdot)$ is FC layer, $W_{OV}^{out}$ is parameter for the FC layer, $h_{OV}^{dec}$ and $c_{OV}^{dec}$ are hidden state vector and output vector in LSTM network, respectively.

## V. PREDICTIVE COLLISION RISK ASSESSMENT

Based on the predicted trajectories of the OVs, the collision risk between the AV and the OVs is assessed with temporal and spatial risk metrics.

*A. Local Trajectory Candidates for the AV*

The longitudinal behaviors of the AV in the prediction horizon can be divided into three types: acceleration, constant speed, and deceleration. The lateral behaviors of the AV can be classified into left lane change, lane-keeping, and right lane change. The combination of longitudinal and lateral behaviors classifies the optional behaviors of the AV in highway scenarios into nine categories. Each behavior type can correspond to multiple candidate trajectories in the trajectory plane.

It is assumed that the longitudinal acceleration of the AV along the longitudinal direction of the road is kept constant in the prediction horizon, the lateral position along the road in the prediction horizon can be expressed by a quintic polynomial curve. The longitudinal position $x_{AV}(t_0)$ and longitudinal velocity $\dot{x}_{AV}(t_0)$ at the initial moment $t_0$ are known, and the longitudinal position of the AV in the predicted horizon can be determined by the longitudinal acceleration $a_x$. The lateral position $y_{AV}(t_0)$, lateral velocity $\dot{y}_{AV}(t_0)$ and lateral acceleration $\ddot{y}_{AV}(t_0)$ at the initial moment $t_0$ are known, and assuming that the lateral velocity $\dot{y}_{AV}(t_0+t_f)$ and lateral acceleration $\ddot{y}_{AV}(t_0+t_f)$ at the end of the predicted horizon are zeros, the quintic polynomial curve characterizing the lateral position can be determined by the lateral position $y_{AV}(t_0+t_f)$ at the end of the predicted horizon.

$$\begin{cases} x_{AV}(t) = x_{AV}(t_0) + \dot{x}_{AV}(t_0)(t-t_0) + \frac{1}{2}a_x(t-t_0)^2 \\ y_{AV}(t) = c_0 + c_1 t + c_2 t^2 + c_3 t^3 + c_4 t^4 + c_5 t^5 \end{cases}, t \in [t_0, t_0+t_f] \quad (14)$$

Where $(x_{AV}, y_{AV})^T$ is the position of the AV in the road curvilinear coordinate system.

As a consequence, the candidate trajectory of the AV in the trajectory plane can be determined by a combination of longitudinal acceleration $a_x$ and lateral position $y_{AV}(t_0+t_f)$. The longitudinal behavior of the AV is represented by the value of the longitudinal acceleration $a_x$. Similarly, the value of the lateral position $y_{AV}(t_0+t_f)$ reflects the lateral behavior. The candidate trajectories corresponding to lane keeping, left lane change and right lane change behaviors for the AV in the prediction horizon are schematically shown in Fig.4.

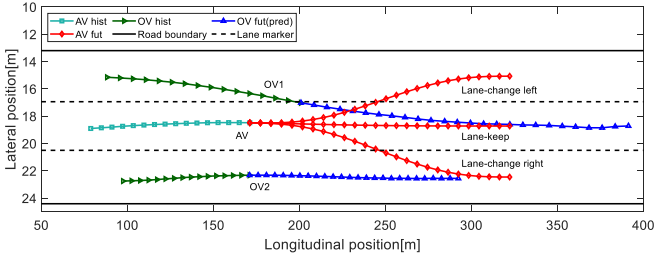

**Fig. 4.** The candidate trajectories of the AV related to three types of lateral behaviors.

*B. Time-continuous Motion Prediction for the AV*

From the vehicle trajectory prediction model, the position of the OV at multiple discrete moments in the predicted horizon can be obtained. The segmented cubic spline curves are used to represent the OV's locations in the predicted horizon. The longitudinal and lateral velocity and acceleration at the initial moment are known. Besides, the interpolation conditions, the conditions of zero-order continuity, first-order continuity, and second-order continuity of the internal nodes are available. Consequently, the coefficients of the segmented cubic spline curves can be determined.

$$x_{OV_i}(t) = \begin{cases} x_{OV_i}^{(1)}(t), t \in [t_0, t_0+\Delta t] \\ x_{OV_i}^{(2)}(t), t \in [t_0+\Delta t, t_0+2\Delta t] \\ \dots \\ x_{OV_i}^{(N_f)}(t), t \in [t_0+t_f-\Delta t, t_0+t_f] \end{cases} \quad (15)$$

$$y_{OV_i}(t) = \begin{cases} y_{OV_i}^{(1)}(t), t \in [t_0, t_0+\Delta t] \\ y_{OV_i}^{(2)}(t), t \in [t_0+\Delta t, t_0+2\Delta t] \\ \dots \\ y_{OV_i}^{(N_f)}(t), t \in [t_0+t_f-\Delta t, t_0+t_f] \end{cases} \quad (16)$$

*C. Risk Metrics Computation*

Traditional collision risk indicators, such as TTC and TH, are mainly based on the assumption that the OV obeys a constant velocity or acceleration model. In addition, there is little consideration given to the vehicle shape. Consequently, the constructed risk indicators are mainly applicable to specific types of collision conflicts. Based on accurate trajectory prediction of the OV, this study uses a combination of temporal and spatial risk metrics to establish a collision risk assessment method applicable to multi-lane highway scenarios, which can assess the collision risk when multiple types of collision conflicts coexist.

*1) Modified TTC Risk Metric:* Similar to the literature [22], separating axis theory [42] was used to determine the collision time between the AV and the OV in the predicted horizon, as illustrated in Fig. 5. The separating axis theory is commonly used for collision detection, taking into account the influence of geometry shapes.

$$TTC_{OV_i} = \begin{cases} \min_{t \in [t_0, t_0+t_f]}\{t-t_0\}, OBB_{AV}(t) \cap OBB_{OV_i}(t) \neq \emptyset \\ t_f, \forall t \in [t_0, t_0+t_f], OBB_{AV}(t) \cap OBB_{OV_i}(t) = \emptyset \end{cases} \quad (17)$$

Where $OBB_{AV}(t)$ and $OBB_{OV_i}(t)$ are oriented bounding box for the AV and the OV, respectively, $TTC_{OV_i}$ is the TTC between the AV and the OV.

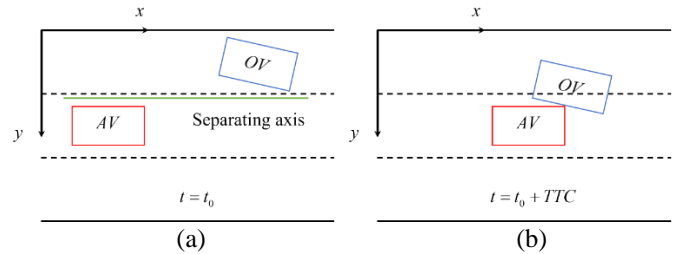

**Fig. 5.** The TTC computation based on separating axis theory: (a) the separating axis exists, there is no collision; (b) the separating axis is non-existent, the collision happens.

*2) The Spatial Risk Metric MDM:* The spacing between the oriented bounding box of the AV and that of the OV along any axis $L$ is as follows.

$$DM_{OV_i}^{(L)}(t) = \max\left\{0, |\boldsymbol{T}\cdot\boldsymbol{L}| - \left[\begin{array}{c}\left|\frac{1}{2}l_{AV}\boldsymbol{e}_{x,AV}\cdot\boldsymbol{L}\right| + \left|\frac{1}{2}w_{AV}\boldsymbol{e}_{y,AV}\cdot\boldsymbol{L}\right| + \\ \left|\frac{1}{2}l_{OV_i}\boldsymbol{e}_{x,OV_i}\cdot\boldsymbol{L}\right| + \left|\frac{1}{2}w_{OV_i}\boldsymbol{e}_{y,OV_i}\cdot\boldsymbol{L}\right|\end{array}\right]\right\}, t\in[t_0,t_0+t_f]$$

(18)

Where $\boldsymbol{T}$ is the vector connecting the center of two oriented bounding boxes, $w_{AV}$, $w_{OV_i}$, $l_{AV}$, and $l_{OV_i}$ are the width and length, respectively, $\boldsymbol{e}_{x,AV}$, $\boldsymbol{e}_{y,AV}$, $\boldsymbol{e}_{x,OV_i}$ and $\boldsymbol{e}_{y,OV_i}$ are the unit vectors along the edge of the oriented bounding boxes.

The minimal distance margin between the two oriented bounding boxes is the minimum of the projected spacing along the four edges.

$$MDM_{OV_i}(t) = \min_{L \in \{e_{x,AV}, e_{y,AV}, e_{x,OV_i}, e_{y,OV_i}\}} \{DM_{OV_i}^{(L)}(t)\} \quad (19)$$

Where $MDM_{OV_i}(t)$ is the MDM between the AV and the OV at the moment $t$.

Further, the minimal distance margin $MDM_{OV_i}(t)$ is decomposed into longitudinal and lateral minimal distance margin, $MDM_{x,OV_i}(t)$ and $MDM_{y,OV_i}(t)$, respectively.

### D. Predictive Collision Risk Assessment between the AV and the OVs

A single risk indicator is often only applicable to a specific collision conflict scenario. The combination of multiple risk indicators can be applied to risk assessment under multiple types of collision conflicts [5, 10]. We formulate collision risk as a time-continuous function in the prediction horizon using temporal and spatial metrics, TTC and MDM. The first item reflects the temporal risk by TTC and time to closest encounter (TTCE) [6]. The second item represent spatial risk related to relative longitudinal and lateral distance.

$$Risk_{OV_i}(t) = w_{TTC} e^{\frac{TTC_{OV_i}^2}{2\sigma_1^2}} e^{-\frac{(t-TTC_{OV_i})^2}{2\sigma_2^2}} + w_{MDM} e^{-\frac{[MSD_{x,OV_i}(t)]^2}{2\sigma_3^2}} e^{-\frac{[MSD_{y,OV_i}(t)]^2}{2\sigma_4^2}}, t \in [t_0, t_0 + t_f] \quad (20)$$

Where $\sigma_1$, $\sigma_2$, $\sigma_\#$, and $\sigma_4$ are model parameters for temporal and spatial risk, $w_{TTC}$ and $w_{MDM}$ are weighting coefficients, $Risk_{OV_i}(t)$ is the collision risk between the AV and the OV at the moment $t$.

Consequently, the collision risk between the AV and all OVs in the local traffic context is obtained [22].

$$Risk(t) = \bigcup_{i=1}^{N_{ov}} Risk_{OV_i}(t) = 1 - \prod_{i=1}^{N_{ov}} \left[1 - Risk_{OV_i}(t)\right] \quad (21)$$

Where $Risk(t)$ is the collision risk between the AV and all OVs at the moment $t$.

## VI. EXPERIMENT AND RESULTS ANALYSIS

In this chapter, the proposed vehicle trajectory prediction model is evaluated on two public highway datasets, and compared with several existing SOTA vehicle trajectory prediction models. Then, simulation results of the predictive collision risk assessment method in typical highway scenarios are analyzed and discussed.

### A. Highway Datasets

The NGSIM US101 [34] and I80 [35] datasets were established in 2005 by US FHWA. The vehicle trajectory was collected by cameras mounted on top of the buildings on the side of the road section. The sampling frequency is 10 Hz. The longitudinal and lateral position labeling accuracy was 4 feet and 2 feet [34-35], respectively. The NGSIM US101 and I80 datasets included a total of 6 data segments. Each segment was 15 minutes. The dataset involved 9206 vehicle trajectories. The NGSIM dataset is currently the most profound public highway dataset in the field of vehicle trajectory prediction.

The highD dataset [36] was created in 2018. A high-definition camera mounted on a drone recorded the vehicle trajectories in six different locations on the German highway. The sampling frequency was 25 Hz. The longitudinal and lateral position labeling accuracy was 0.1m [36]. The dataset consisted of 60 segments, each about 17 minutes long, recording a total of about 110,000 vehicles traveling through the observed road section.

### B. Quantitative Evaluation of Vehicle Trajectory Prediction Model

*1) Data Pre-processing:* High data quality is the key to ensure accurate prediction of vehicle trajectory with a data-driven approach. However, the public available NGSIM dataset involves noise data. The first-order Butterworth low-pass filter is used to filter the position, velocity, and acceleration time series data in the NGSIM US101 and I80 datasets. The cut-off frequency is set to 1 Hz. The position, velocity, and acceleration time series data in the highD dataset have been processed by Rauch-Tung-Striebe (RTS) smoothing method [36] and can be used directly in subsequent studies.

*2) Model Implementation Details:* An 8s time window is used to sample the vehicle trajectory passing the observed road section. The first 3s is the historical trajectory and the last 5s is the future trajectory. The time interval is $\Delta t = 0.2s$. The sliding window moves at a time step of 1s. The NGSIM dataset and highD dataset are divided into the training set, validation set and test set according to 70%, 10%, and 20%, respectively. The model parameters in the vehicle trajectory prediction model are optimized with Adam algorithm. The batch size is 256. The learning rate is 0.001. The cycles of pre-training and formal training are set to 5 and 10, respectively. The early-stopping strategy is used to prevent model parameters from overfitting. The vehicle trajectory prediction model is implemented based on PyTorch deep learning architecture. The computing platform is a high-performance workstation configured with Intel Xeon Gold 6139 2.30GHz CPU and NVIDIA GeForce RTX 3090 24GB GPU.

*3) Quantitative Comparisons with Existing SOTA methods:* Using the RMSE of the predicted trajectory as a criterion, the proposed trajectory prediction model is evaluated on the NGSIM and highD datasets, and compared with existing SOTA methods. The prediction results of the compared methods on both datasets are taken from the published works. A brief introduction of the compared methods is as follows.

*CS-LSTM [17]:* Convolutional social pooling operation is used to extract local spatial vehicle interactions.

*NLS-LSTM [38]:* Non-local social pooling is introduced to model distant vehicle interactions.

*MHA-LSTM(+f) [18]:* Multi-head attention mechanism is adopted to represent global vehicle interactions. In addition to position features, velocity, acceleration, and vehicle type features are involved.

*Structural-LSTM [19]:* Based on radial spatial connections, nearby vehicles share the hidden state vector and output vector in LSTM networks.

*GNN-RNN [39]:* RNN is used to extract each vehicle's temporal relationship. GNN is adopted to model spatial interactions among vehicles.

*EA-Net [40]:* Graph attention network and convolutional social pooling are combined for modeling the interactions between the OV and the local traffic context.

*DRM-DL [41]:* Driving risk map is utilized to represent the interactions between the OV and the driving environment.

*HEAT-R [43]:* Three channels encoder the historical motion of the OV, the interactions among traffic participants, and map structure information, respectively.

The quantitative comparison between the prosed CSP-GAN-LSTM model and existing SOTA models is illustrated in Tab. I. The proposed vehicle trajectory model outperforms the existing SOTA models over the whole prediction horizon (1-5 s) on the NGSIM dataset. Meanwhile, on the highD dataset, the proposed CSP-GAN-LSTM model has an equal prediction accuracy with SOTA models in the short-term (1-2 s), and outperforms the existing models in the long-term (3-5 s). Compared with the existing SOTA methods, the RMSEs of the CSP-GAN-LSTM at 5s on the NGSIM and highD datasets have reduced by 9% and 51%, respectively.

In addition, the prediction accuracy of the proposed CSP-GAN-LSTM model on the highD dataset is much better than that on the NGISM dataset. The different prediction accuracy can be caused by the discrepancy in data scale and position labeling accuracy. The number of vehicle tracks in the highD dataset is 12 times higher than in the NGSIM dataset, which ensures that more vehicle interaction configurations are involved. Besides, the position labelling accuracy is 0.1m in the highD dataset, which is more accurate than that of the NGSIM dataset. As a result, the performance of the trajectory prediction model can be optimized in terms of both data annotation accuracy and data scale.

TABLE I
QUANTITATIVE PREDICTION RESULTS ON TWO HIGHWAY DATASETS OVER A 5S PREDICTION HORIZON (RMSE IN METERS)

| Models | NGSIM US 101 and I80 datasets | | | | | highD dataset | | | | |
| --- | --- | --- | --- | --- | --- | --- | --- | --- | --- | --- |
| | Prediction horizon | | | | | Prediction horizon | | | | |
| | 1s | 2s | 3s | 4s | 5s | 1s | 2s | 3s | 4s | 5s |
| CS-LSTM [IEEE CVPR 2018] | 0.61 | 1.27 | 2.09 | 3.10 | 4.37 | 0.22 | 0.61 | 1.24 | 2.10 | 3.27 |
| NLS-LSTM [IEEE IV 2019] | 0.56 | 1.22 | 2.02 | 3.03 | 4.30 | 0.20 | 0.57 | 1.14 | 1.90 | 2.91 |
| MHA-LSTM(+f) [IEEE T-IV 2021] | **0.41** | 1.01 | 1.74 | 2.67 | 3.83 | **0.06** | **0.09** | **0.24** | 0.59 | 1.18 |
| Structural-LSTM [IEEE T-ITS 2020] | 0.60 | 1.09 | 1.38 | 1.68 | **1.95** | 0.29 | 0.49 | 0.65 | 0.82 | 1.16 |
| GNN-RNN [IEEE ITSC 2021] | 0.68 | 0.99 | 1.21 | 1.53 | 2.14 | - | - | - | - | - |
| EA-Net [IEEE Trans. VT 2021] | 0.42 | **0.88** | 1.43 | 2.15 | 3.07 | 0.15 | 0.26 | 0.43 | 0.78 | 1.32 |
| DRM-DL [IEEE T-ITS 2022] | - | - | - | - | - | 0.17 | 0.29 | 0.38 | **0.58** | **0.93** |
| HEAT-R [IEEE T-ITS 2022] | 0.68 | 0.92 | **1.15** | **1.45** | 2.05 | - | - | - | - | - |
| CSP-GAN-LSTM (This work) | **0.19** | **0.48** | **0.82** | **1.23** | **1.78** | **0.05** | **0.09** | **0.17** | **0.26** | **0.46** |

TABLE II
ABLATION ANALYSIS FOR THE PROPOSED CSP-GAN-LSTM MODEL (RMSE IN METERS)

| Models | NGSIM US 101 and I80 datasets | | | | | highD dataset | | | | |
| --- | --- | --- | --- | --- | --- | --- | --- | --- | --- | --- |
| | Prediction horizon | | | | | Prediction horizon | | | | |
| | 1s | 2s | 3s | 4s | 5s | 1s | 2s | 3s | 4s | 5s |
| Channel 1 | 0.35 | 1.12 | 2.19 | 3.55 | 5.18 | 0.07 | 0.12 | 0.30 | 0.74 | 1.55 |
| Channel1 + Channel 2 | **0.19** | 0.49 | 0.84 | 1.27 | 1.86 | **0.04** | **0.06** | **0.11** | **0.22** | 0.58 |
| Channel 1 +Channel 3 | 0.21 | 0.52 | 0.86 | 1.31 | 1.93 | 0.08 | 0.11 | 0.22 | 0.30 | 0.71 |
| Channel 1 + Channel 2 + Channel 3 | **0.19** | **0.48** | **0.82** | **1.23** | **1.78** | 0.05 | 0.09 | 0.17 | 0.26 | **0.46** |

*4) Model Ablation Analysis:* To verify the effectiveness of the three-channel feature fusion operation, an ablation analysis is implemented on the proposed CSP-GAN-LSTM model.

*Channel 1:* Only the historical motion state sequence of the OV is used as input.

*Channel 1 + Channel 2:* The input contains the historical track of the OV and the SVs. Besides Channel 1, the vehicle interactions in the local context are modeled with convolutional social pooling operation.

*Channel 1 + Channel 3:* In addition to Channel 1, the interactions between the OV and the SVs is represented with a graph attention network.

*Channel 1 + Channel 2 + Channel 3:* The complete CSP-GAN-LSTM model proposed in this work. The features from three channels are fused for decoding the future motion of the OV.

The model ablation results are illustrated in Tab. II. Ignoring the vehicle interactions, the prediction accuracy of the *Channel 1* is the lowest. The introduction of vehicle interactions makes the prediction accuracy of *Channel 1 + Channel 2* and *Channel 1 + Channel 3* outperform that of *Channel 1*. Furthermore, the combination of different types of vehicle interactions ensures excellent prediction performance in the long-term (5s).

*5) Effects of Input Feature Types:* To investigate the effect of input feature types on trajectory prediction results, the input feature type settings are shown in Tab. III. The prediction results with different input feature types are illustrated in Fig. 6. With only position information, the CSP-GAN-LSTM model has the lowest prediction accuracy. The introduction of velocity and acceleration information contributes to the improvement of model prediction accuracy. In addition, the introduction of the relative motion information between the OV and the SVs greatly improves the model prediction accuracy.

TABLE III
DIFFERENT INPUT FEATURE TYPES

| Input feature types | | OV | SV |
|---|---|---|---|
| Position information | Absolute motion | $\boldsymbol{x}_{OV}(t)=\left[x_{ov}(t),y_{ov}(t)\right]^{T}$ | $\boldsymbol{x}_{SV}(t)=\left[x_{SV_i}(t),y_{SV_i}(t)\right]^{T}$ |
| | Absolute and relative motion | $\boldsymbol{x}_{OV}(t)=\left[x_{ov}(t),y_{ov}(t)\right]^{T}$ | $\boldsymbol{x}_{SV}(t)=\left[x_{SV_i}(t),y_{SV_i}(t),\Delta x_{SV_i}(t),\Delta y_{SV_i}(t)\right]^{T}$ |
| Position and velocity information | Absolute motion | $\boldsymbol{x}_{OV}(t)=\begin{bmatrix}x_{ov}(t),y_{ov}(t),\\ \dot{x}_{ov}(t),\dot{y}_{ov}(t)\end{bmatrix}^{T}$ | $\boldsymbol{x}_{SV}(t)=\left[x_{SV_i}(t),y_{SV_i}(t),\dot{x}_{SV_i}(t),\dot{y}_{SV_i}(t)\right]^{T}$ |
| | Absolute and relative motion | $\boldsymbol{x}_{OV}(t)=\begin{bmatrix}x_{ov}(t),y_{ov}(t),\\ \dot{x}_{ov}(t),\dot{y}_{ov}(t)\end{bmatrix}^{T}$ | $\boldsymbol{x}_{SV}(t)=\begin{bmatrix}x_{SV_i}(t),y_{SV_i}(t),\dot{x}_{SV_i}(t),\dot{y}_{SV_i}(t),\\ \Delta x_{SV_i}(t),\Delta y_{SV_i}(t),\Delta\dot{x}_{SV_i}(t),\Delta\dot{y}_{SV_i}(t)\end{bmatrix}^{T}$ |
| Position, velocity, and acceleration information | Absolute motion | $\boldsymbol{x}_{OV}(t)=\begin{bmatrix}x_{ov}(t),y_{ov}(t),\dot{x}_{ov}(t),\\ \dot{y}_{ov}(t),\ddot{x}_{ov}(t),\ddot{y}_{ov}(t)\end{bmatrix}^{T}$ | $\boldsymbol{x}_{SV}(t)=\left[x_{SV_i}(t),y_{SV_i}(t),\dot{x}_{SV_i}(t),\dot{y}_{SV_i}(t),\ddot{x}_{SV_i}(t),\ddot{y}_{SV_i}(t)\right]^{T}$ |
| | Absolute and relative motion | $\boldsymbol{x}_{OV}(t)=\begin{bmatrix}x_{ov}(t),y_{ov}(t),\dot{x}_{ov}(t),\\ \dot{y}_{ov}(t),\ddot{x}_{ov}(t),\ddot{y}_{ov}(t)\end{bmatrix}^{T}$ | $\boldsymbol{x}_{SV}(t)=\begin{bmatrix}x_{SV_i}(t),y_{SV_i}(t),\dot{x}_{SV_i}(t),\dot{y}_{SV_i}(t),\ddot{x}_{SV_i}(t),\ddot{y}_{SV_i}(t),\\ \Delta x_{SV_i}(t),\Delta y_{SV_i}(t),\Delta\dot{x}_{SV_i}(t),\Delta\dot{y}_{SV_i}(t),\Delta\ddot{x}_{SV_i}(t),\Delta\ddot{y}_{SV_i}(t)\end{bmatrix}^{T}$ |

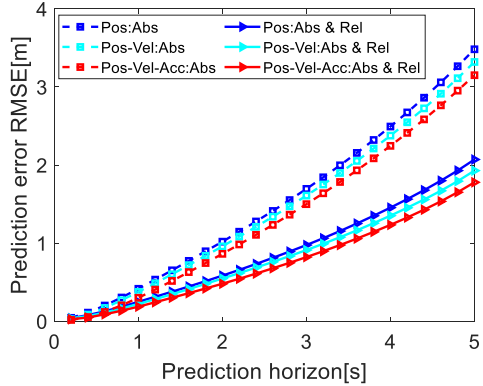
(a)

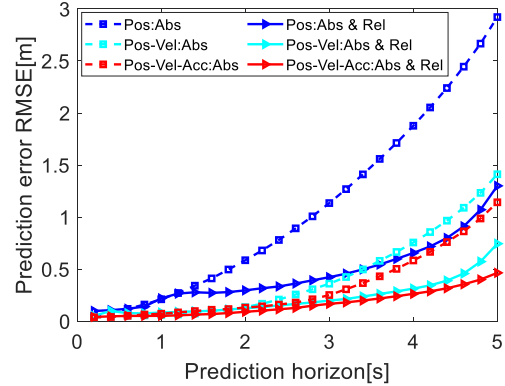
(b)

**Fig. 6.** Effects of Input Feature Types: (a) NGSIM dataset; (b) highD dataset.

*C. Results Discussion and Analysis in Typical Highway Scenarios*

Typical highway scenarios are selected from the German highway dataset highD to validate the feasibility and effectiveness of the proposed predictive collision risk assessment method. The road structure is two-way four lanes, as illustrated in Fig. 7. The parameters for vehicle shape, road structure, and risk assessment are defined in Tab. IV. We mainly focus on predictive risk assessment for typical lane-keeping and lane change behaviors, and the the expected lateral position $y_{AV}(t_0+t_f)$ at the end of the prediction horizon equals $y_{Lane1}$ or $y_{Lane2}$.

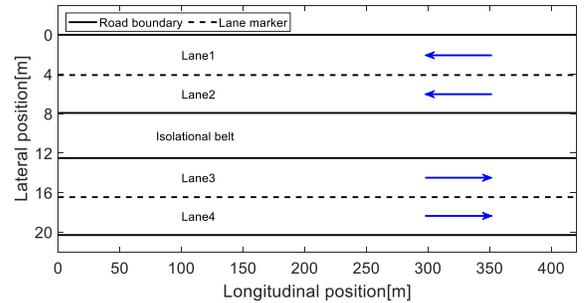

**Fig. 7.** Road structure.

TABLE IV
MODEL PARAMETERS

| Symbol | Description | Value | Unit |
|---|---|---|---|
| $l_{AV}, l_{OV_i}$ | Vehicle length | 5.21 | m |
| $w_{AV}, w_{OV_i}$ | Vehicle width | 2.04 | m |
| $a_x$ | Longitudinal acceleration | [-5,5] | m/s$^2$ |
| $y_{Lane1}$ | Lateral position of lane 1 | 2 | m |
| $y_{Lane2}$ | Lateral position of lane 2 | 6 | m |
| $y_{Lane3}$ | Lateral position of lane 3 | 14.4 | m |
| $y_{Lane4}$ | Lateral position of lane 4 | 18.4 | m |
| $\sigma_1$ | Parameter for temporal risk related to TTC | 2.04 | s |
| $\sigma_2$ | Parameter for temporal risk related to TTCE | 2.04 | $\sigma_1$ |
| $\sigma_3$ | Parameter for longitudinal spatial risk | 45 | m |
| $\sigma_4$ | Parameter for lateral spatial risk | 1.6 | m |
| $w_{TTC}$ | Weight for temporal risk | 0.6 | - |
| $w_{MDM}$ | Weight for spatial risk | 0.4 | - |

*1) Car-following Scenario:* In highway scenario 1, the AV and OVs are located in lane 1 and lane 2, traveling along the negative direction of the $x$ axis. The AV and OVs make lane-keeping behaviors. The AV follows the preceding $OV1$ at a similar speed. At the same time, $OV2$ is moving at high speed in front of the left adjacent lane.

Fig. 8 and Fig. 9 illustrate the predictive collision risk assessment results for the AV at the moment $t_0 = 4s$ and $t_0 = 6s$, respectively. The longitudinal velocity in the historical horizon of the AV and OVs is shown in Fig. 8(a) and Fig. 9(a). The trajectory prediction results of the OVs are depicted in Fig. 8(b) and Fig. 9(b). At the moment $t_0 = 4s$, the maximal longitudinal and lateral position errors in the whole prediction horizon are 0.33 m and 0.07m, respectively. At the moment $t_0 = 6s$, the maximal longitudinal and lateral position errors in the whole prediction horizon is 0.55 m and 0.07 m.

The longitudinal distance between the AV and the $OV2$ increases over time. Consequently, the collision risk between the AV and $OV2$ decreases. The main collision risk for AV is resulted from $OV1$. The collision risk maps in Fig. 8(d) and Fig. 9(d) indicate that AV's longitudinal acceleration behavior will cause high collision risk. In order to mitigate the collision risk, the optional behaviors for the AV are keeping its current lane at constant speed or deceleration, or making a lane change left to the adjacent lane.

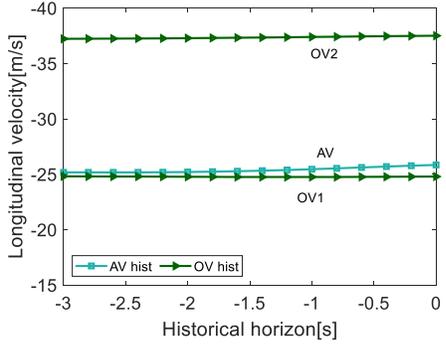

(a)

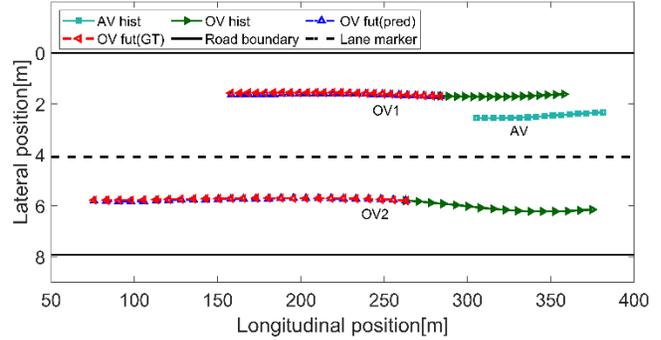

(b)

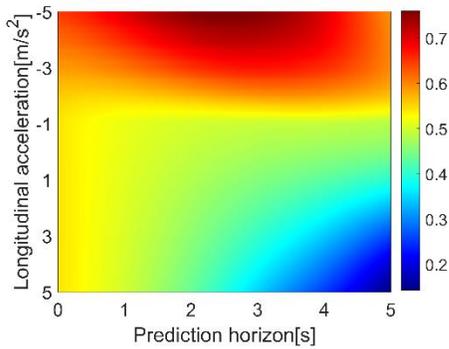

(c)

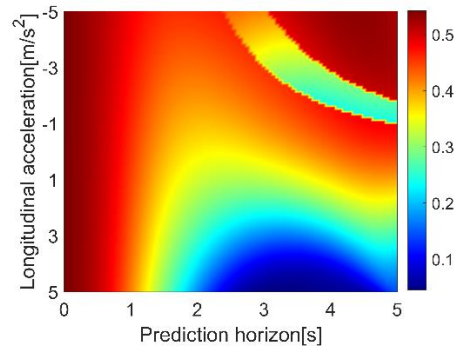

(d)

**Fig. 8.** Predictive collision risk assessment results at $t_0 = 4s$ in car-following scenario: (a) Longitudinal velocity of the AV and OVs in the historical horizon; (b) Predicted trajectories of the OVs; (c) Predictive collision risk under lane-keeping behavior; (d) Predictive collision risk under left lane-change behavior.

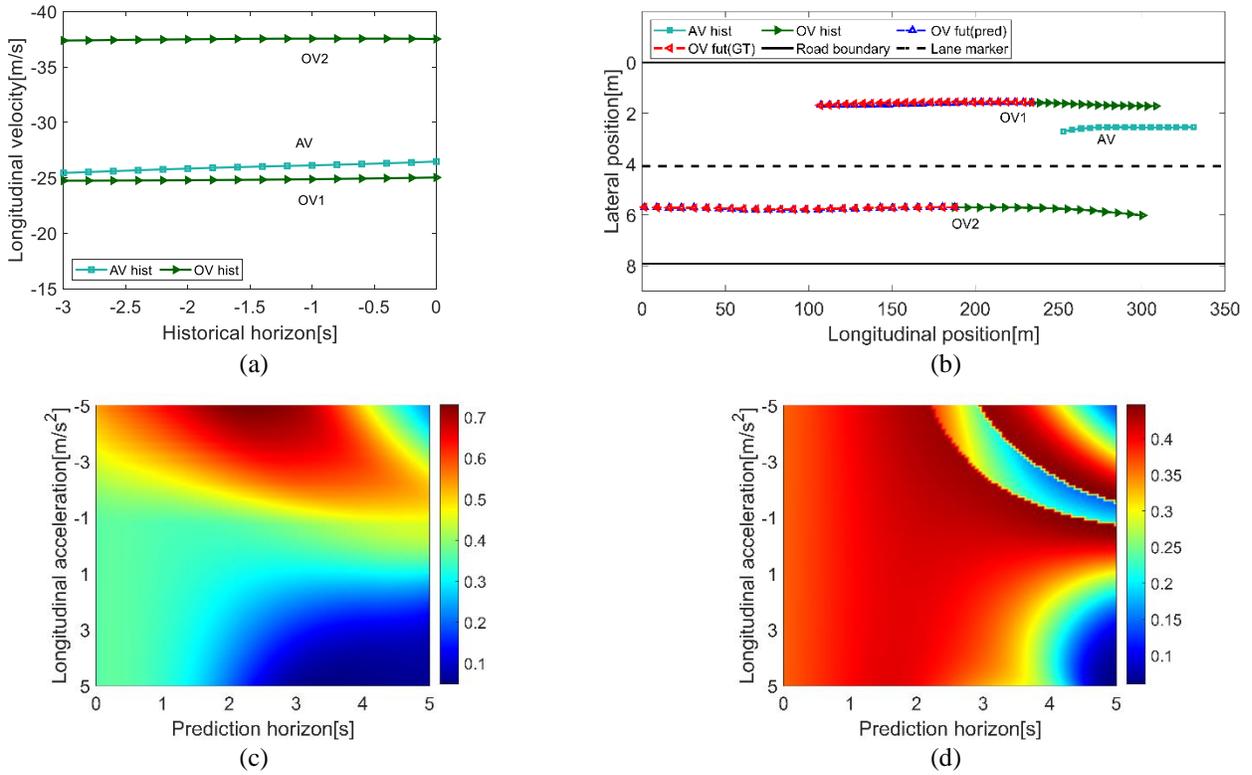

**Fig. 9.** Predictive collision risk assessment results at $t_0 = 6s$ in car-following scenario: (a) Longitudinal velocity of the AV and OVs in the historical horizon; (b) Predicted trajectories of the OVs; (c) Predictive collision risk under lane-keeping behavior; (d) Predictive collision risk under left lane-change behavior.

*2) Preceding Vehicle in Adjacent Lane Cut-in Scenario:* In highway scenario 2, the AV is making a left lane change. The $OV1$ and $OV3$ keep their original lane, while $OV2$ is cutting into the area in front of the AV.

At the moment $t_0 = 5s$ and $t_0 = 7s$, the trajectory prediction results of $OV1$, $OV2$, and $OV3$ is shown in Fig. 10(b) and Fig. 11(b). The predicted trajectory of OVs coincides approximately with the ground truth. The maximal longitudinal and lateral position prediction errors at $t_0 = 5s$ are 1.33 m and 0.08m. At the moment $t_0 = 7s$, the maximal longitudinal and lateral position prediction errors are 0.88 m and 0.12 m, respectively. The slightly large longitudinal position prediction error is caused by the lane change behavior of $OV2$.

The predictive collision risk assessment results at $t_0 = 5s$ and $t_0 = 7s$ are shown in Fig. 10(c), Fig. 10(d), Fig. 11(c), and Fig. 11(d). Obviously, accelerated lane-keeping or lane change behavior will result in high collision risk. Consequently, in order to avoid high collision risk, the optional longitudinal behavior for AV in the prediction horizon is deceleration or constant velocity.

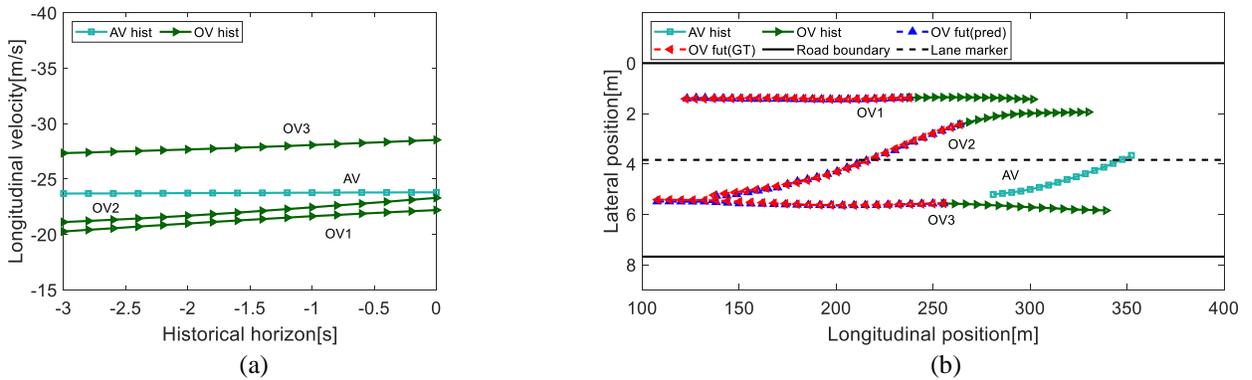

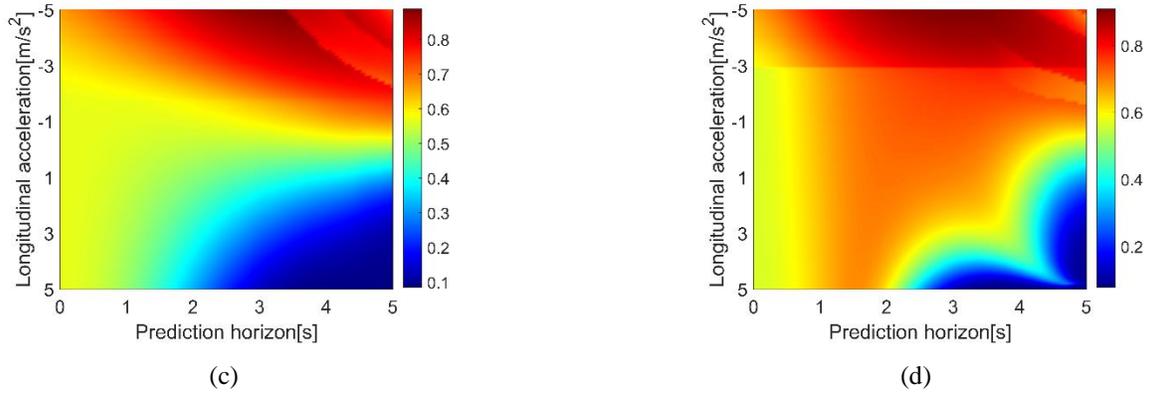

**Fig. 10.** Predictive collision risk assessment results at $t_0 = 5s$ in cut-in scenario: (a) Longitudinal velocity of the AV and OVs in the historical horizon; (b) Predicted trajectories of the OVs; (c) Predictive collision risk under lane-keeping behavior; (d) Predictive collision risk under right lane-change behavior.

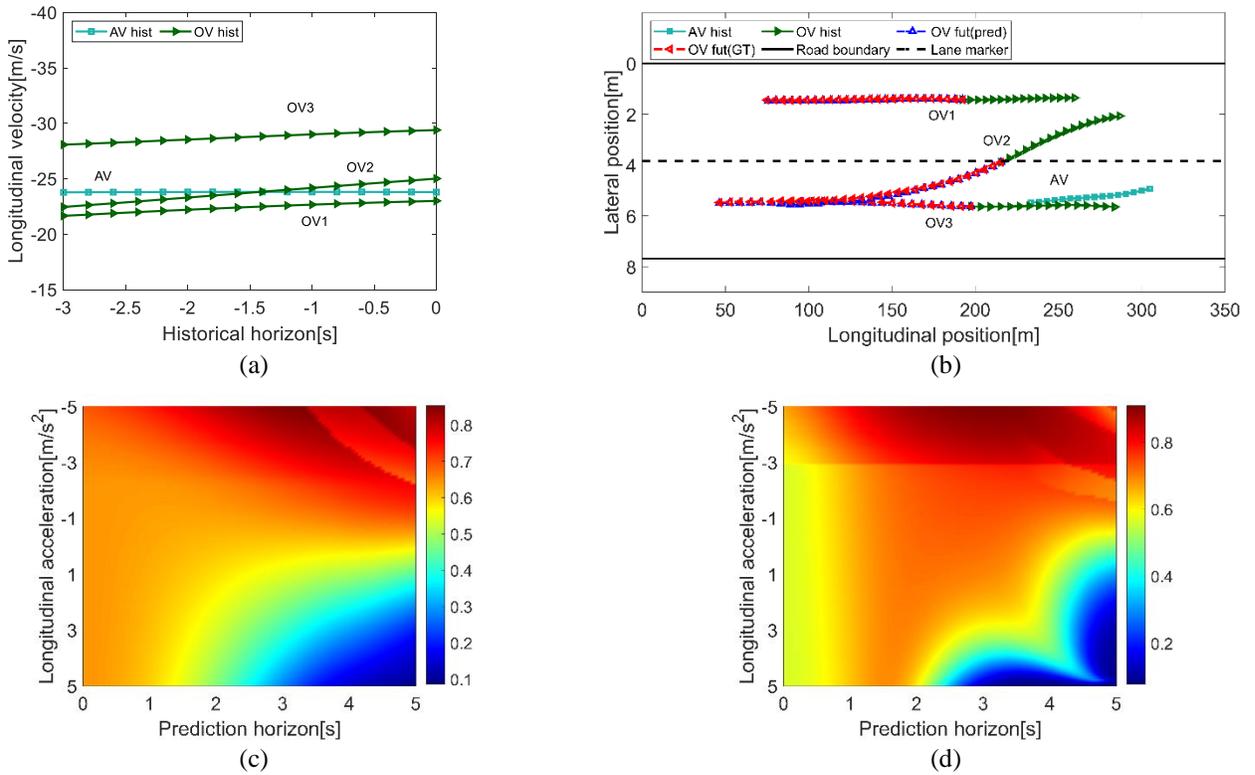

**Fig. 11.** Predictive collision risk assessment results at $t_0 = 7s$ in cut-in scenario: (a) Longitudinal velocity of the AV and OVs in the historical horizon; (b) Predicted trajectories of the OVs; (c) Predictive collision risk under lane-keeping behavior; (d) Predictive collision risk under right lane-change behavior.

## VII. SUMMARY AND CONCLUSIONS

For autonomous driving in highway scenarios, a predictive collision risk assessment method based on trajectory prediction of the OVs is proposed. Three channels are adopted to extract historical motion temporal relationships of the OV, the local spatial vehicle interactions among vehicles in the local traffic context, and global vehicle interactions between the OV and its SVs, respectively. The proposed CSP-GAN-LSTM vehicle trajectory prediction model is evaluated on two public highway datasets. The quantitative results indicate that the CSP-GAN-LSTM model outperforms the existing SOTA models in terms of prediction accuracy. Then, based on the predicted trajectories of the OVs and the candidate trajectory of the AV, the predictive collision risk between the AV and the OVs is assessed by the combination of temporal and spatial risk metrics. The simulation results in typical highway scenarios validate the feasibility and effectiveness of the proposed predictive collision risk assessment method.

Despite the excellent performance, the proposed CSP-GAN-LSTM model belongs to a multi-input-single-output architecture. There is insufficient consideration of the vehicle interactions in the predicted horizon. In future works, the multi-vehicle trajectory prediction task should be established with a multi-input-multi-output framework, which incorporates vehicle interactions in the future. In addition, the proposed predictive collision risk assessment method is only suitable for highway scenarios. In the future, predictive collision risk assessment in urban scenarios should be explored based on simultaneously predicting the trajectory of multiple types of traffic participants.

REFERENCES

[1] R. Behringer, S. Sundareswaran, B. Gregory, R. Elsley, B. Addison, W. Guthmiller, R. Daily, D. Bevly. The DARPA grand challenge-development of an autonomous vehicle. In *Proc. IEEE Intell. Veh. Symp.*, 2004: 226-231.
[2] M. Buehler, K. Iagnemma, and S. Singh. *The 2005 DARPA grand challenge-the great robot race*. Berlin, German: Springer, 2007.
[3] M. Buehler, K. Iagnemma, and S. Singh. *The DARPA urban challenge-autonomous vehicles in city traffic*. Berlin, German: Springer, 2009.
[4] L. Claussmann, M. Revilloud, D. Gruyer, and S. Glaser. A review of motion planning for highway autonomous driving. *IEEE Trans. Intell. Transp. Syst.* vol 21, no. 5, pp. 1826-1848, May 2020.
[5] C. Xu, W. Zhao, and C. Wang. An integrated threat assessment algorithm for decision-making of autonomous driving vehicles. *IEEE Trans. Intell. Transp. Syst.* vol. 21, no. 6, pp. 2510-2521, Jul. 2019.
[6] F. Damerow, and J. Eggert. Predictive risk map. In *proc. IEEE 17th Int. Conf. Intell. Transp. Syst.* 2014, pp. 703-710.
[7] S. Lefèvre, D. Vasquez, and C. Laugier. A survey on motion prediction and risk assessment for intelligent vehicles. *ROBOMECH J.* vol. 1, no. 1, pp. 1-14, Jul. 2014.
[8] U.Z.A. Hamid, Y. Saito, H. Zamzuri, M.A.A. Rahman, and P. Raksinchroensak. A review on threat assessment, path planning and path tracking strategies for collision avoidance systems of autonomous vehicles. *Int. J. Veh. Auton. Syst.* vol. 14, no. 2, pp. 134-169, Nov. 2018.
[9] J. Dahl, G. R. D. Campos, C. Olsson, and J. Fredriksson. Collision avoidance: a literature review on threat-assessment techniques. *IEEE Trans. Intell. Veh.* vol. 4, no. 1, pp. 101-113, Mar. 2019.
[10] Y. Wang, C. Wang, W. Zhao, and C. Xu. Decision-making and planning method for autonomous vehicles based on motivation and risk assessment. *IEEE Trans. Veh. Technol.* vol. 70, no. 1, pp. 107-120, Jan. 2021.
[11] G. Li, Y. Yang, T. Zhang, X. Qu, D. Cao, B. Cheng, and K. Li. Risk assessment based collision avoidance decision-making for autonomous vehicles in multi-scenarios. *Transp. Res. Part C: Emerg. Technol.* vol 122, 102820, pp. 1-17, Jan. 2021.
[12] A. Polychronopoulos, M. Tsogas, A. J. Amditis, and L. Andreone. Sensor fusion for predicting vehicles' path for collision avoidance systems. *IEEE Trans. Intell. Transp. Syst.* vol. 8, no. 3, pp. 549-562, Sep. 2007.
[13] A. Houenou, P. Bonnifait, V. Cherfaoui, and W. Yao. Vehicle trajectory prediction based on motion model and maneuver recognition. In *Proc. IEEE Int. Conf. Intell. Robot. Syst.*, pp. 4363-4369, 2013.
[14] G. Xie, H. Gao, L. Qian, B. Huang, K. Li, and J. Wang. Vehicle trajectory prediction by integrating physics- and maneuver-based approaches using interactive multiple models. *IEEE Trans. Ind. Electron.* vol. 65, no. 7, pp. 5999-6008, Jul. 2018.
[15] M. Schreier, V. Willert, and J. Adamy. An integrated approach to maneuver-based trajectory prediction and criticality assessment in arbitrary road environments. *IEEE Trans. Intell. Transp. Syst.* vol. 17, no. 10, pp. 2751-2766, Oct. 2016.
[16] S. Mozaffari, O. Y. Al-Jarrah, M. Dianati, P. Jennings, and A. Mouzakitis. Deep learning-based vehicle behavior prediction for autonomous driving applications: a review. *IEEE Trans. Intell. Transp. Syst.* vol. 23, no. 1, pp. 33-47, Jan. 2022.
[17] N. Deo, and M. M. Trivedi. Convolutional social pooling for vehicle trajectory prediction. In *Proc. IEEE Conf. Comp. Vis. Pattern Reconi.*, pp. 1549-1557, 2018.
[18] K. Messaoud, I. Yahiaoui, A. V. Blondet, and F. Nashashibi. Attention based vehicle trajectory prediction. *IEEE Intell. Veh.*, vol. 6, no. 1, pp. 175-185, Mar. 2021.
[19] L. Hou, L. Xin, S. E. Li, B. Cheng, and W. Wang. Interactive trajectory prediction of surrounding road users for autonomous driving using structural-LSTM network. *IEEE Trans. Intell. Transp. Syst.* vol. 21, no. 11, pp. 4615-4625, Nov. 2020.
[20] X. Li, X. Ying, and M. C. Chuah. GRIP: graph-based interaction-aware trajectory prediction. In *Proc. IEEE Intell. Transp. Syst. Conf.*, pp. 3960-3966, 2019.
[21] R. Chandra, T. Guan, S. Panuganti, T. Mittal, U. Bhattacharya, A. Bera, and D. Manocha. Forecasting trajectory and behavior of road-agents using spectral clustering in graph-LSTMs. *IEEE Robot. Automat. Lett.*, vol. 5, no. 3, pp. 4882-4890, Jul. 2020.
[22] K. Kim, and D. Kum. Collision risk assessment algorithm via lane-based probabilistic motion prediction of surrounding vehicles. *IEEE Trans. Intell. Transp. Syst.* vol. 19, no. 9, pp. 2965-2976, Sep. 2018.
[23] J. Hayward. *Near misses as a measure of safety at urban intersections*, University Park, PA, The Pennsylvania State University, 1971.
[24] F. Raymond. Determinants of time headway adopted by truck drivers. *Ergonomics*. vol. 24, no. 6, pp. 463-474, 1981.
[25] S. Mahmud, L. Ferrira, M. S. Hoque, and A. Tavassoli. Application of proximal surrogate indicators for safety evaluation: a review of recent developments and research needs. *Int. Assoc. Traffic Saf. Sci. Res*. vol. 41, no. 4, pp. 153-163, Dec. 2017.
[26] S. Noh S, and W. Y. Han. Collision avoidance in on-road environment for autonomous driving. In *Proc. Int. Conf. Control, Automat. Syst*, 2014, pp. 884-889.
[27] Y. Kuang, Y. Yu, and X. Qu. Novel crash surrogate measure for freeways. *J. Transp. Engineering, Part A: Syst.*, vol. 146, no. 8, pp. 1-10, 2020.
[28] S. Tak, S. Kim, D. Lee, and H. Yeo. A comparison analysis of surrogate safety measures with car-following perspective for advanced driver assistance system. *J. Adv. Transp.*, Paper ID 8040815, pp. 1-14, 2018.
[29] L. Wang L, C. F. Lopez, and C. Stiller. Realistic single-shot and long-term collision risk for a human-style safer driving. In *Proc. IEEE Intell. Symp.*, 2020, pp. 2073-2080.
[30] X. Shi, Y. D. Wong, C. Chai, M. Z. F. Li, T. Chen, and Z. Zeng. Automatic clustering for unsupervised risk diagnosis of vehicle driving for smart road. 2020, arXiv:2011.11933.
[31] M. Althoff, and A. Mergel. Comparison of Markov chain abstraction and Monte Carlo simulation for the safety assessment of autonomous cars. *IEEE Trans. Intell. Transp. Syst.*, vol. 12, no.4, pp. 1237-1247, Dec. 2011.
[32] M. Althoff, O. Stursberg, and M. Buss. Model-based probabilistic collision detection in autonomous driving. *IEEE Trans. Intell. Transp. Syst.*, vol. 10, no. 2, pp. 299-310, Jun. 2009.
[33] A. Lawitzky, D. Althoff, C. F.Passenberg, G. Tanzmeister, D. Wollherr, and M. Buss. Interactive scene prediction for automotive applications. In Proc. *IEEE Intell. Veh. Symp.*, pp. 1028-1033, 2013.
[34] J. Colyar and J. Halkias. *US highway 101 dataset*. In Federal Highway Administration (FHWA). Tech. Rep. FHWA-HRT-07-030, 2007.
[35] J. Colyar and J. Halkias. *US highway i-80 dataset*. In Federal Highway Administration (FHWA). Tech. Rep. FHWA-HRT-06-137, 2006.
[36] R. Krajewski, J. Bock, L. Kloeker, and L. Eckstein. The highD dataset: a drone dataset of naturalistic vehicle trajectories on German highway for validation of highly autonomous driving systems. *21st Int. Conf. Intell. Transp. Syst.* 2018, pp. 2118-2125.
[37] M. Montanino and V. Punzo. Trajectory data reconstruction and simulation-based validation against macroscopic traffic patterns. *Transp. Res. Part B.* vol. 80, pp. 82-106, 2016.
[38] K. Messaoud, I. Yahiaoui, A. V. Blonder, and F. Nashashibi. Non-local social pooling for vehicle trajectory prediction. In *Proc. IEEE Intell. Veh. Symp.* 2019, pp. 975-980.
[39] X. Mo, Y. Xing, and C. Lv. Graph and recurrent neural network-based vehicle trajectory prediction for highway driving. In *Proc. IEEE Intell. Transp. Syst. Conf.* 2021, pp. 1934-1939.
[40] Y. Cai, Z. Wang, H. Wang, L. Chen, Y. Li, M. A. Sotelo, and Z. Li. Environment-attention network for vehicle trajectory prediction. *IEEE Trans. Veh. Technol.* vol. 70, no. 11, pp. 11216-11217, Nov. 2021.
[41] X. Liu, Y. Wang, K. Jiang, Z. Zhou, K. Nam, and C. Yin. Interactive trajectory prediction using a driving risk map-integrated deep learning method for surrounding vehicles on highways. *IEEE Trans. Intell. Transp. Syst.* Early Access Mar. 30, 2022, doi: 10.1109/TITS.2022.3160630.
[42] J. Huynh. (2009). Separating Axis Theorem for Oriented Bounding Boxes. [Online]. Available: www.jkh.me/files/.
[43] X. Mo, Z. Huang, Y. Xing, and C. Lv. Multi-agent trajectory prediction with heterogeneous edge-enhanced graph attention network. *IEEE Intell. Transp. Syst.* Early Access. Feb. 1, 2022, doi: 10.1109/TITS.2022.3146300.